%% file: 2018_EvoNetGenesNIPS.tex
\documentclass{article}

\PassOptionsToPackage{numbers, compress}{natbib}

\usepackage[final]{nips_2018}

\usepackage[utf8]{inputenc} 
\usepackage[T1]{fontenc}    
\usepackage{hyperref}       
\usepackage{url}            
\usepackage{booktabs}       
\usepackage{amsfonts}       
\usepackage{nicefrac}       
\usepackage{microtype}      
\usepackage[cmex10]{amsmath}
\usepackage{amssymb}
\usepackage{graphicx}

\title{Mitigating Architectural Mismatch During the Evolutionary Synthesis of Deep Neural Networks}

\author{
  Audrey G. Chung$^{*}$, Paul Fieguth, Alexander Wong\\
  Vision and Image Processing Research Group, University of Waterloo, Waterloo, ON, Canada\\
  Waterloo Artificial Intelligence Institute, University of Waterloo, Waterloo, ON, Canada\\
  \texttt{$^{*}$ agchung@uwaterloo.ca} \\
}

\begin{document}

\maketitle

\begin{abstract}
\input{./Abstract.tex}
\end{abstract}
\section{Introduction}
\input{./Introduction.tex}
\section{Methods}
\input{./Methods.tex}

\section{Results}
\input{./Results.tex}
\section{Discussion}
\vspace{-10pt}
\input{./Discussion.tex}
\section*{Acknowledgements}
\vspace{-5pt}
This research has been supported by the Canada Research Chairs Program and Natural Sciences and Engineering Research Council of Canada (NSERC). The authors also thank Nvidia for the GPU hardware used in this study through the Nvidia Hardware Grant Program.


\end{document}

%% file: Abstract.tex
\textit{Evolutionary deep intelligence} has recently shown great promise for producing small, powerful deep neural network models via the organic synthesis of increasingly efficient architectures over successive generations. Existing evolutionary synthesis processes, however, have allowed the mating of parent networks independent of architectural alignment, resulting in a mismatch of network structures. We present a preliminary study into the effects of architectural alignment during evolutionary synthesis using a gene tagging system. Surprisingly, the network architectures synthesized using the gene tagging approach resulted in slower decreases in performance accuracy and storage size; however, the resultant networks were comparable in size and performance accuracy to the non-gene tagging networks. Furthermore, we speculate that there is a noticeable decrease in network variability for networks synthesized with gene tagging, indicating that enforcing a like-with-like mating policy potentially restricts the exploration of the search space of possible network architectures.

%% file: Introduction.tex
Deep neural networks (DNNs)~\cite{LeCun2015, Bengio2009, Graves2013, Tompson2014} have recently garnered widespread interest due to their demonstrated ability to improve state-of-the-art performance in many challenging areas of research. This boost in performance, however, has largely been attributed to increasingly large and complex network architectures, resulting in storage and memory requirements beyond the resources available in practical scenarios. As such, methods for producing highly efficient DNNs have been developed to reduce the memory and computational needs while maintaining performance accuracy.

Inspired by nature, Shafiee~\textit{et al.}~\cite{Shafiee2016} introduced \textit{evolutionary deep intelligence} as a biologically-motivated alternative to compressing existing DNNs directly by allowing networks to organically synthesize new and increasingly efficient network architectures over successive generations. While existing evolutionary deep intelligence methods have leveraged asexual evolutionary synthesis~\cite{Shafiee2016,Shafiee2016_2,Shafiee2017,Shafiee2017_2,Shafiee2017_3} and $m$-parent sexual evolutionary synthesis~\cite{Chung2017,Chung2017_2,Chung2018}, the evolutionary synthesis process combines architectural structures sequentially regardless of their relative positions in each parent network. As a result, current evolutionary deep intelligence methods have employed evolutionary synthesis processes that combine parent networks independent of architectural alignment, resulting in a mismatch of network structures.

In this paper, we present a preliminary study into the effects of architectural alignment during evolutionary synthesis via the introduction of a gene tagging system. Gene tagging is explored within the context of $m$-parent sexual evolutionary synthesis and evaluated over a range of environmental resource models. The gene tagging system allows for the proper alignment of architectural structures that originated from the same location in the ancestor network, and enforces a like-with-like mating policy during evolutionary synthesis. 

%% file: Methods.tex
In this study, we modify the existing $m$-parent evolutionary synthesis approach~\cite{Chung2017_2} to investigate the efficacy of structural alignment and its effect on the synthesis of new network architectures across a range of simulated environmental resource models.

\subsection{$m$-Parent Evolutionary Synthesis}
Let the network architecture be formulated as $\mathcal{H}(N,S)$, where $N$ is the set of possible neurons and $S$ denotes the set of possible synapses in the network. Each neuron $n_j \in N$ is connected to neuron $n_k \in N$ via a set of synapses $\bar{s} \subset S$, such that the synaptic connectivity $s_j \in S$ has an associated $w_j \in W$ to denote the connection's strength. In the seminal evolutionary deep intelligence paper~\cite{Shafiee2016}, the synthesis probability $P(\mathcal{H}_{g}|\mathcal{H}_{g-1}, \mathcal{R}_{g})$ of a new network at generation $g$ is approximated by the synaptic probability $P(S_g|W_{g-1}, R_{g})$ to emulate heredity through the generations of networks. $P(\mathcal{H}_{g}|\mathcal{H}_{g-1}, \mathcal{R}_{g})$ is also conditional on an environmental factor model $\mathcal{R}_g$ to imitate natural selection via simulated environmental resources. 

Generalizing to $m$-parent evolutionary synthesis~\cite{Chung2017_2}, a newly synthesized network $\mathcal{H}_{g(i)}$ can be dependent on a subset of all previously synthesized networks $\mathcal{H}_{G_i}$, where the set of network indices $G_i$ corresponds to the set of previous networks on which $\mathcal{H}_{g(i)}$ is dependent, and $g(i)$ represents the generation number corresponding to the $i^{\text{th}}$ network. Note that in the general case, the number of networks in subset $\mathcal{H}_{G_i}$ and the range of generational dependency $g(G_i)$ is only constrained by the number and generational range of already synthesized networks.

The synthesis probability combining the probabilities of $m$ parent networks $\mathcal{H}_{G_i}$ is represented by some cluster-level mating function $\mathcal{M}_c(\cdot)$ and some synapse-level mating function $\mathcal{M}_s(\cdot)$:
\begin{align}
P(\mathcal{H}_{g(i)}|\mathcal{H}_{G_i}, \mathcal{R}_{g(i)}) =& \prod_{C \in \mathcal{C}} \Big[ P(s_{g(i),C}|\mathcal{M}_c(W_{\mathcal{H}_{G_i}}), \mathcal{R}_{g(i)}^c) \cdot \nonumber \\ &\prod_{j \in C} P(s_{g(i),j}|\mathcal{M}_s(w_{\mathcal{H}_{G_i},j}), \mathcal{R}_{g(i)}^s) \Big].
\end{align} 
These mating functions are formulated using an intersection-based mating policy~\cite{Chung2017_2}, i.e.,:
\begin{align}
	\mathcal{M}_c(W_{\mathcal{H}_{G_i}}) = \prod_{k = 1}^{m} \alpha_{c,k} W_{\mathcal{H}_k} \hspace{70pt}
	\mathcal{M}_s(w_{\mathcal{H}_{G_i},j}) = \prod_{k = 1}^{m}\alpha_{s,k} w_{\mathcal{H}_k,j}
\end{align}
where $W_{\mathcal{H}_k}$ represents the cluster's synaptic strength for the $k^{th}$ parent network $\mathcal{H}_k \in \mathcal{H}_{G_i}$. Similarly, $w_{\mathcal{H}_k,j}$ represents the synaptic strength of a synapse $j$ within a given cluster for the $k^{th}$ parent network $\mathcal{H}_k \in \mathcal{H}_{G_i}$. 

\subsection{Mitigating Architecture Mismatch via Gene Tagging}
To encourage like-with-like mating during evolutionary synthesis, this study introduces a gene tagging system to enforce structural alignment, i.e., only mating architectural clusters originating from the same location in the ancestor network. As such, the cluster-level and synapse-level mating functions proposed in~\cite{Chung2017_2} are reformulated as follows:
\begin{align}
	\mathcal{M}_c(\overline{W}_{\mathcal{H}_{G_i}}) = \prod_{k \in \mathcal{K}_c} \alpha_{c,k} \overline{W}_{\mathcal{H}_k} \hspace{70pt}	\mathcal{M}_s(\overline{w}_{\mathcal{H}_{G_i},j}) = \prod_{k \in \mathcal{K}_c}\alpha_{s,k} \overline{w}_{\mathcal{H}_k,j}
\end{align}
where $\mathcal{K}_c$ is the subset of parent networks with existing architectural clusters corresponding to a single gene tagged cluster $c \in C$, $C$ is the set of clusters that exists in $\mathcal{H}_{g(i)}$, and $\overline{W}$ and $\overline{w}$ are the gene tagged and structurally-aligned synaptic strengths. 

This study also introduces a parameter-based lower bound on the subset of parent networks with existing architectural clusters $\mathcal{K}_c$. Prior studies~\cite{Chung2017, Chung2017_2,Chung2018} employed an intersection-based mating policy where a cluster was synthesized only if the cluster existed within all $m$ parent networks (i.e., $\mathcal{K}_c = m$). To allow for more flexibility within the $m$-parent evolutionary synthesis, we propose a percent of population parameter $\chi$ to control the proportion of parent network architectures that must contain any given cluster during evolutionary synthesis. As such, only clusters that exist within the specified proportion of parent networks are synthesized (i.e., $m\chi \leq \mathcal{K}_c \leq m$), and the intersection-based policy can be achieved using $\chi = 1$.

%% file: Results.tex
\subsection{Experimental Setup}
The effect of structural alignment during the $m$-parent evolutionary synthesis was evaluated using 10\% of the MNIST~\cite{LeCun1998} hand-written digits dataset to increase the training speed of the synthesized network architectures and increase the inherently low intra-class variation within the MNIST dataset. For this preliminary study, 5-parent evolutionary synthesis and a $\chi$ of $60\%$ was used, and $\mathcal{H}_{G_i}$ was restricted to networks in the immediately preceding generation, i.e., for a newly synthesized network $\mathcal{H}_{g(i)}$, the $m = 5$ parent networks in $\mathcal{H}_{G_i}$ are from generation $g(i)-1$.

The first generation ancestor network was trained using the LeNet-5 architecture~\cite{LeCun1998_LeNet}, and the synthesized offspring networks were assessed using performance accuracy on the MNIST dataset and storage size (representative of the architectural efficiency of a network) of the networks with respect to the computational time required. Similar to~\cite{Chung2018}, each filter was considered as a synaptic cluster in the multi-factor synapse probability model, and the cluster-level environmental factor model $\mathcal{R}^c_{g(i)}$ and the synapse-level environmental factor model $\mathcal{R}^s_{g(i)}$ were varied together from 50\% to 95\% at 5\% increments, i.e.,:
\begin{align}
	\mathcal{R}^c_{g(i)}, \mathcal{R}^s_{g(i)} = \{50, 55, 60, 65, 70, 75, 80, 85, 90, 95\} \%,
\end{align}
to simulate the varying availability of environmental resources during $m$-parent evolutionary synthesis. The synthesized networks were assessed using performance accuracy on the MNIST dataset and storage size (representative of the architectural efficiency of a network) of the networks with respect to the computational time required.

\subsection{Experimental Results}

\begin{figure*}[t]
\centering
\begin{tabular}{cc}
\includegraphics[width=.45\textwidth]{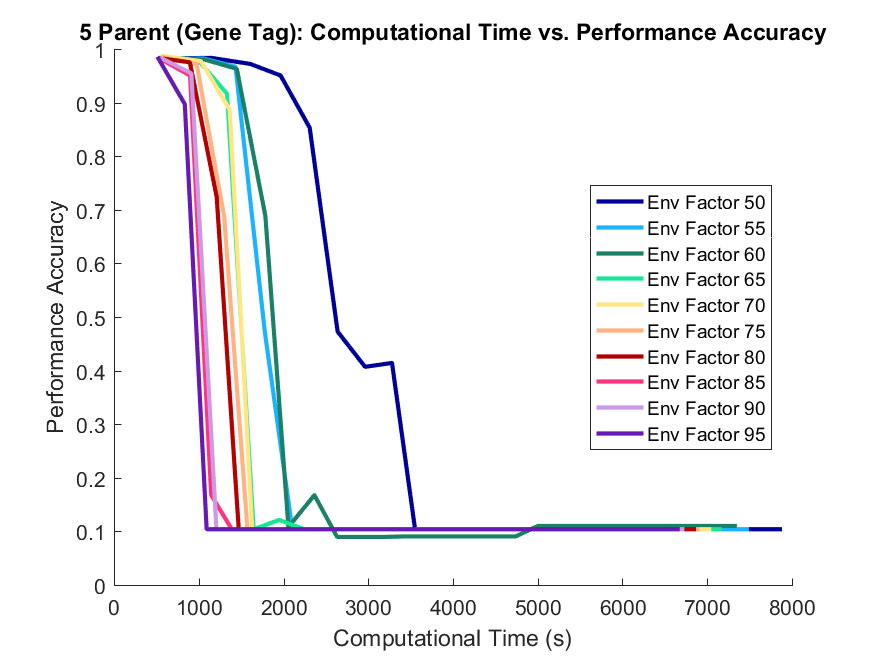}&
\includegraphics[width=.45\textwidth]{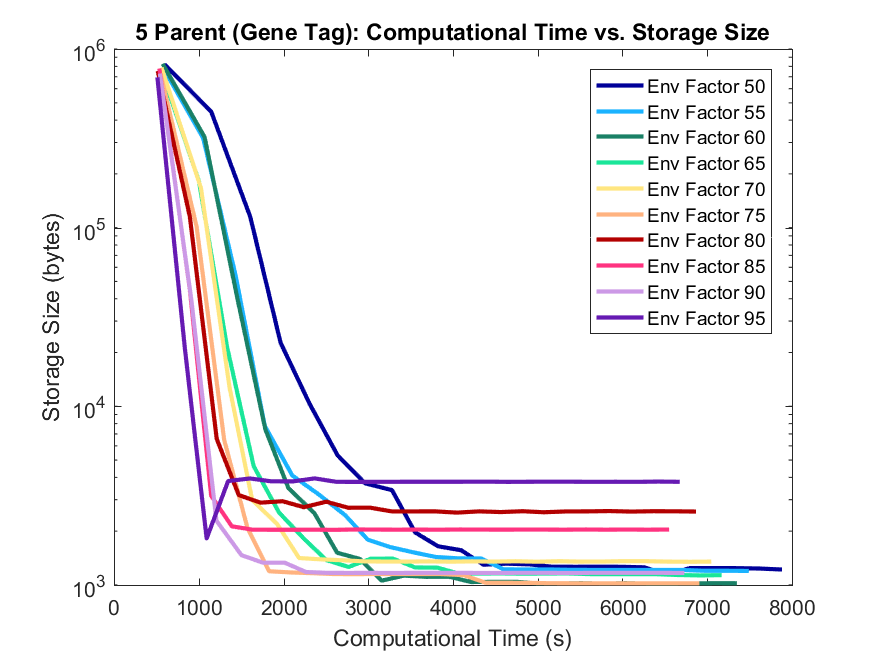}\\
\includegraphics[width=.45\textwidth]{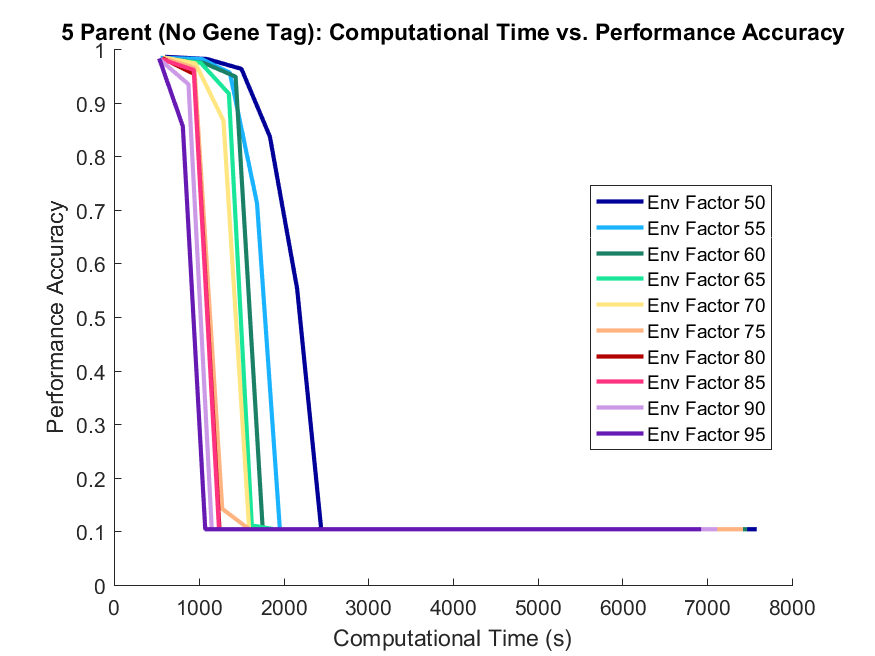}&
\includegraphics[width=.45\textwidth]{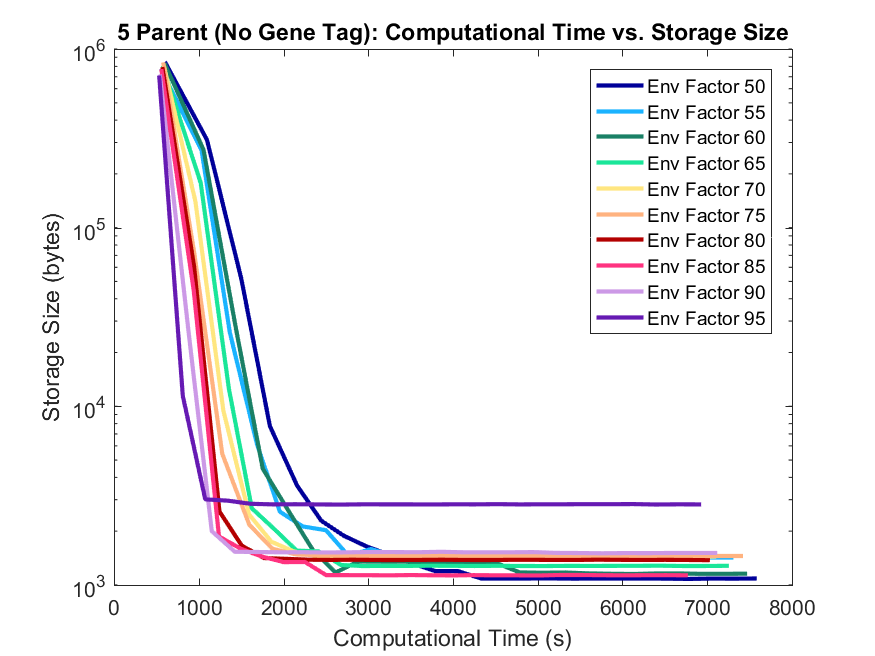}\\
\end{tabular}
\caption{Performance accuracy (left) and storage size (right) for 5-parent evolutionary synthesis with gene tagging (top row) and no gene tagging (bottom row) using various environmental factor models. Plots best viewed in colour.}
\label{fig_Results}
\end{figure*}

Figure~\ref{fig_Results} shows the performance accuracy and storage size for 5-parent evolutionary synthesis with gene tagging (top row) and no gene tagging (bottom row) given various cluster-level and synapse-level environmental factor models. The incorporation of gene tagging resulted in a more gradual decrease in both performance accuracy and storage size relative to computational time, while the omission of gene tagging produced network architectures with more rapidly decreasing performance accuracy and storage size. Note that the bottom plateau in performance accuracy is at $10\%$ akin to random guessing, as the MNIST dataset consists of 10 classes of handwritten digits. While there is no inherent limit in network reduction when using the evolutionary deep intelligence approach, Figure~\ref{fig_Results} shows that there are natural plateaus in network storage size, particularly with the higher environmental factor models, i.e., environmental factor models of $80\%$, $85\%$, and $95\%$. In addition, note that these plateaus in network storage size tend to occur more when synthesizing networks using gene tagging.

\begin{figure}[t]
\centering
\includegraphics[width=0.65\linewidth]{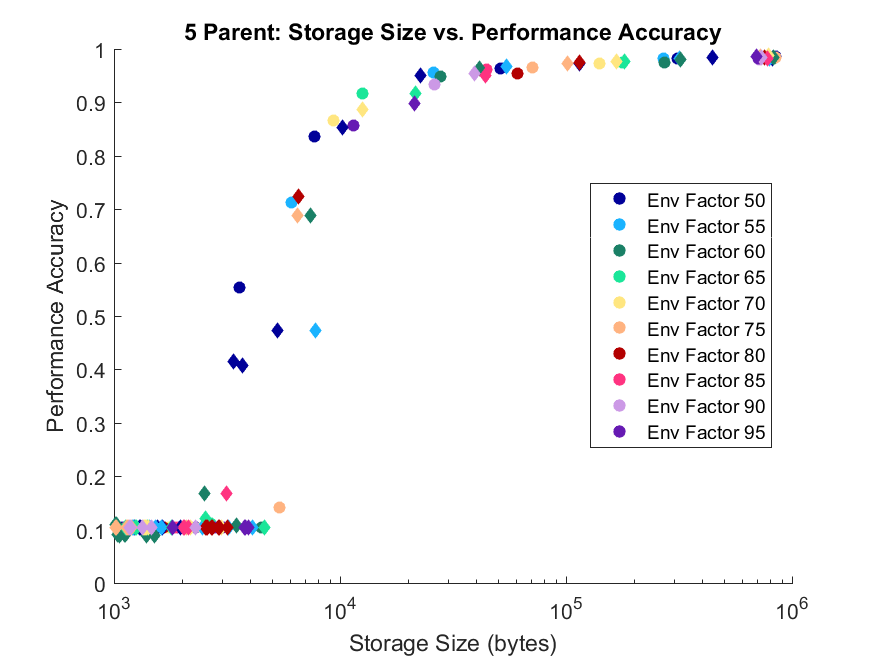}
\caption{Performance accuracy as a function of storage size for 5-parent sexual evolutionary synthesis using various environmental factor models. Networks synthesized using gene tagging (diamond) show minimal to no noticeable difference relative to networks synthesized without gene tagging (round) in terms of maintaining performance accuracy while decreasing storage size. Plots best viewed in colour.}
\label{fig_ResultsScatter}
\end{figure}

Figure~\ref{fig_ResultsScatter} shows performance accuracy as a function of storage size for 5-parent sexual evolutionary synthesis using various cluster-level and synapse-level environmental factor models, where the best synthesized networks are closest to the top left corner, i.e., high performance accuracy and low storage size. Networks synthesized using gene tagging (diamond points) appear to be minimally worse relative to networks synthesized without gene tagging (round points) in terms of maintaining performance accuracy while decreasing storage size.

%% file: Discussion.tex
In this work, we presented a preliminary study into the effects of architectural alignment during evolutionary synthesis via the introduction of a gene tagging system to enforce a like-with-like mating policy.
We speculate that the use of gene tagging inherently decreases overall network variability across the synthesized network architectures, indicating that enforcing a like-with-like mating policy via the use of gene tagging potentially restricts the exploration of the search space of possible network architectures. As a result, synthesized networks may fail to achieve the same increases in architectural efficiency as architectures synthesized with more random approaches. Future work includes further investigation into the effects of architectural alignment via the use of different $\chi$s and varying the number of parents networks $m$, and the exploration of how gene tagging potentially affects the architectural variability in synthesized networks. 